# Software Vulnerability Detection via Deep Learning over Disaggregated Code Graph Representation

Yufan Zhuang, Sahil Suneja, Veronika Thost, Giacomo Domeniconi, Alessandro Morari, Jim Laredo

*IBM Research, Yorktown Heights, NY, USA*

## Abstract

Identifying vulnerable code is a precautionary measure to counter software security breaches. Tedious expert effort has been spent to build static analyzers, yet insecure patterns are barely fully enumerated. This work explores a deep learning approach to automatically learn the insecure patterns from code corpora. Because code naturally admits graph structures with parsing, we develop a novel graph neural network (GNN) to exploit both the semantic context and structural regularity of a program, in order to improve prediction performance. Compared with a generic GNN, our enhancements include a synthesis of multiple representations learned from the several parsed graphs of a program, and a new training loss metric that leverages the fine granularity of labeling. Our model outperforms multiple text, image and graph-based approaches, across two real-world datasets.

## 1 Introduction

Recently, AI research has made inroads in source code understanding, and being able to perform tasks such as program invariant inferencing [33], code categorization [56], code variable prediction [11], and function naming [10], amongst others. Similarly, in case of software vulnerability detection, what used to be a domain traditionally dominated by static and dynamic analysis is seeing assistance and competition from AI models. Shortcomings in existing techniques, such as the high false positives of static analyzers, and the lack of completeness of dynamic analysis, are a few reasons promoting the entry of AI into this field [52, 62].

In particular, Graph Neural Networks [48] are surfacing as the state of the art in AI-based vulnerability detection. This is because of their ability to operate directly on graph-structured input, combined with the graph-level representation that can be mapped naturally on to source code. These approaches typically operate on source code converted to compiler-level graph structures, incorporating the syntactic and semantic relationships between program statements via abstract syntax tree (AST), data flow graph (DFG), and control flow graph (CFG).

Previous work [61] demonstrated that manually-created graph-level vulnerability templates are useful in searching for similar instances elsewhere in a project. With source code represented as a graph, a template can be defined as a control-flow and data-dependency relationship between locations of interest in source code (i.e. edge connectivity between nodes of a graph). For example, the absence of a 'variable sanitization template' [61], i.e. value-range validation prior to being used as a memory allocation size argument, may indicate a potential buffer overflow vulnerability. In this work, we explore the possibility to automatically learn such vulnerability templates by analyzing multiple programs, and then match them across unseen source code graphs to catch vulnerabilities in them. In a machine learning setting, the vulnerability detection problem can be translated to a supervised classification task. This boils down to training a model over several examples of source code snippets, labeled as being vulnerable or healthy, with the goal of getting the model to extract vulnerability patterns or signals from them.

In this paper, we present a novel graph neural network (GNN) architecture designed to assist the model in concentrating on specific program constructs independently. Unlike existing GNN-based approaches which are based on a single multi-graph (fusing one or more graphs together) [19, 62], our model operates on different representations of the source code graph ({AST, CFG, DFG}) separately. By focusing on the graphs individually, the model is afforded more opportunity to better learn vulnerable patterns which each graph captures. Then, the aggregated information for each graph is only composed at the end, when forming the representation of the entire program for the downstream classification task.

Our model, which we call 3GNN, is based on Crystal Graph Convolutional Networks (CGCN) [60] and Self Attention Graph (SAG) Pooling [31]. As shown in figure 1, the three source code graph components are used to obtain initial vector representations. These are used as input for three separate pipelines with shared weights, which are learned by the model while being trained across multiple examples of vulnerable and healthy source code. These vector representations are in turn the input of multiple hierarchical modules comprised of CGCN and SAG Pooling layers. Each layer drops irrelevant graph nodes to assist the next layer to narrow down towards the set of nodes and edges better representative of vulnerabilities. Finally, the three pipelines converge into a multi-layer perceptron, which performs the final classification of input code into vulnerable or healthy, based upon the relative probability the model prescribes the input code to each class. Along with a novel model architecture, we also present a new loss function specifically designed for the vulnerability detection task, to measure (and improve upon) how far off the model prediction is for the input source code sample. We evaluate our model on two publicly available C/C++ datasets: Draper [36], and QEMU+FFmpeg [62]. Our results show that 3GNN achieves superior performance compared to various text (BiLSTM), image (CNN) and graph-based (GGNN) baselines, recording a 6.9% better F1 than the next-best model (GGNN).

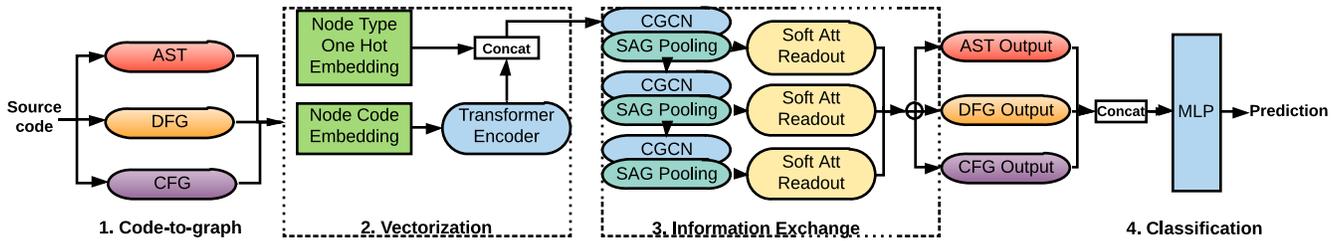

**Figure 1:** Our 3GNN model architecture with its 4 stages. Input source code is first transformed into AST, CFG and DFG graphs to feed to the model. Each graph is processed in three independent pipelines. Each pipeline follows the same Vectorization and Information Exchange steps. The final vector representations across the three pipelines are aggregated together, and fed to the final classification MLP layer, which generates the prediction for the sample in terms of relative probability of a sample being vulnerable or healthy.

## 2 Motivation

**Traditional Approaches:** Traditionally, vulnerability detection has been grouped into the categories of static analyses and dynamic analyses [20]. Static analysis tools, such as Clang static analyzer [2] and Infer [4], typically build a model of program states and reason over all possible behaviors that might happen in the real execution. Constrained by a vast search space of possible executions and behaviors, static analyses make trade-offs in favor of scalability and abstract some details, lose information and hence produce False Positives (code tagged as being vulnerable but isn't so). For rule-based static analyses, such as linters [1] and taint analysis tools [3], the quality of the results depend on the coverage of the defect types and the quality of the rules. By contrast, dynamic analyses execute the program and observe the execution behaviors. Testing, symbolic execution, concolic testing and fuzzing [5–8, 15, 51] are commonly used for this purpose. While it can concretely expose the defects in the execution, it requires a precise input that can drive the execution to the places of interests. Unfortunately, it's usually challenging to prepare inputs that achieve either good program coverages or satisfying the particular path conditions.

**AI Augmentation Approach:** In contrast, with AI-based approaches, the aim is to learn such code-to-vulnerability mapping heuristics automatically. The idea is pretty straightforward–presenting a large dataset of tagged examples to a learning-based model so that it can figure out the properties which differentiate vulnerable from healthy code. For example, when a model is exposed to enough examples (and counter-examples) like the one shown in Figure 2(a), it may learn that the existence of an `if` construct is an important signal in the context of avoiding buffer overflow vulnerabilities. Similarly, by being exposed (trained) to examples of other vulnerability types, the model can be taught to pick up the corresponding signals or templates [61], such as for pointer dereference, resource leak, use-after-free, and failure to release lock, amongst others. The learned heuristics from such a macroscopic approach to vulnerability detection can then augment the microscopic approach of static analyzers. The end result is a more finely curated set of alerts to the developer to aid in secure code development with hopefully lesser False Positives.

The promise of this has been shown in previous work, with learning-based approaches outperforming static analyzers in vulnerability detection. For example, [45,52] report F1[1] in the range of 0.02 to 0.45 for different static analyzers, compared to 0.84 for a neural network which treats source code simply as photos. These results are for the NIST Juliet test suite [40], developed specifically for measuring the effectiveness of static analysis tools.

**Our Approach:** Unlike existing approaches which treat source code as images [27], or as a sequence of tokens [35,42], we focus on the graph nature of source code which we feel is a more natural representation of code, encapsulating semantically rich information. Unlike existing graph neural network approaches [19, 62] which are based on a single graph component, or a fused multi-graph, our model operates on the AST, CFG, and DFG separately. Each graph excels at exposing particular vulnerabilities through concentrating on the respective program constructs while omitting irrelevant information. For example, the CFG pairing `malloc`s and `free`s detects potential resource leaks in program branches. Hence, when constructing vector representations of a program by using a graph neural network, it is more beneficial to perform graph convolution and message passing on each graph separately, maximally exploiting vulnerable patterns a specialized graph captures. This way, information of a node in the single multi-graph, but irrelevant to a particular sub-graph targeting specific vulnerability types, will not be aggregated blindly for this sub-graph, enhancing its opportunity of exposing the vulnerability. Aggregated information for each graph will only be composed at the end, when forming the representation of the entire program for the downstream classification task. Furthermore, with bigger program size, which is quite often the case with real-world code, the corresponding graphs can become quite big. This makes it harder for the model to efficiently gather information across the graph, during learning. Treating the graph components separately limits this information flow complexity, assisting in more efficient learning.

---
[1] F1 is the harmonic mean of Precision and Recall.

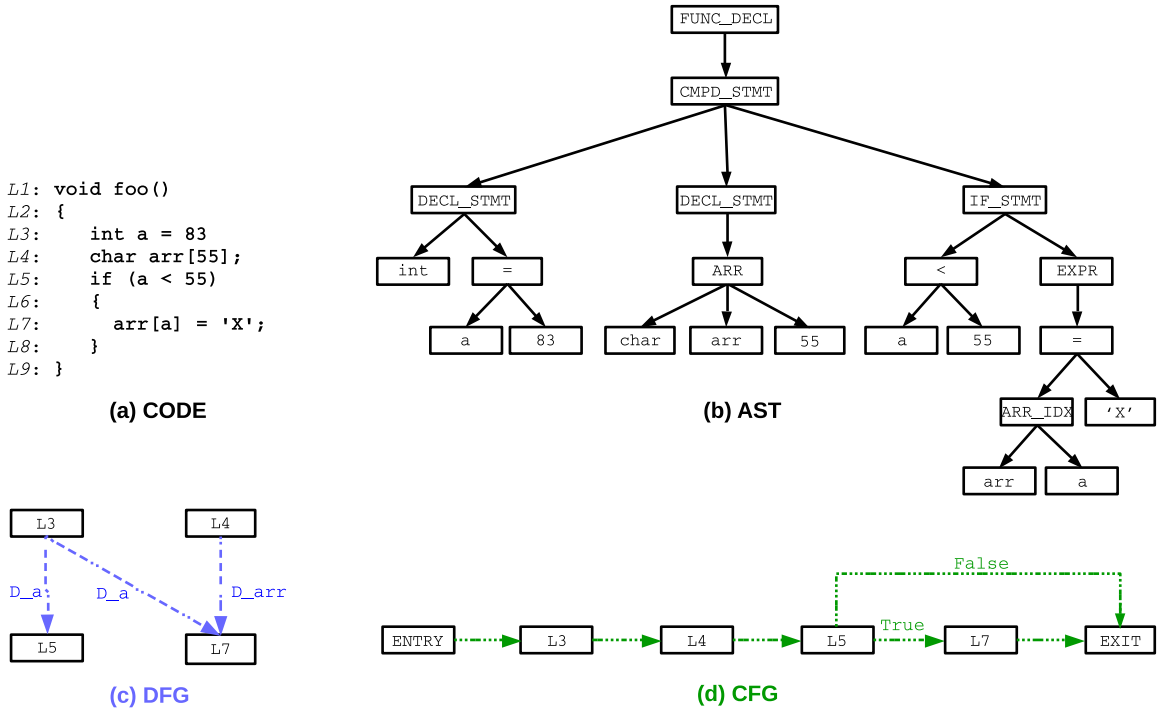

**Figure 2:** A sample code snippet with its corresponding AST, CFG and DFG graphs.

## 3 Design

Figure 1 shows our 3GNN model architecture, which takes in raw source code and classifies it as being vulnerable or not. It consists of the following 4 stages: (i) converting code to graphs, (ii) encoding graphs as vectors, (iii) propagating information through the graphs, and (iv) classifying final vectors as representing vulnerable or healthy code.

### 3.1 Code-to-Graph Transformation

The first step is the conversion of source code into a representation amenable for consumption by neural networks. With graph neural networks, which can operate on graph inputs directly, we are able to preserve syntactic and semantic relationships of source code components, by using compiler-level graph structures to represent code. In particular, the code's syntax information is captured by the AST, as shown in Figure 2(b). The AST models the structures of the function but is insufficient to capture the program behavior. Semantic information is encapsulated by the data flow (DFG) and control flow (CFG) graphs. In Figure 2(c), the *Use-Def* edges with tag "D_a" represent the data dependency from the variable `a` definition in line 3, to it's use in lines 5 and 7 of the code snippet shown in Figure 2(a). Similarly, the edges of the CFG (Figure 2(d)) capture the execution order such as conditional branches. We use the Joern open-source tool [21] to convert each code sample into its graph representation.

### 3.2 Vectorization

After converting source code to graphs, the next step is to convert its nodes and edges into a vector representation consumable by neural networks.

**Node Vectors:** Each node in the code graph is composed of a node type and its source code component. While using Joern to convert source code to graph, we get 69 node types, for example *Identifier*, *CallExpression*, *IfStatement* etc. For each node, the type is encoded into a one-hot vector[2]. Each node's source code component, on the other hand, is encoded through a transformer encoder [55], which preserves semantic information between the source code tokens. We select transformer [55] as the encoder to encode the source code since it has been shown to provide better contextualized embeddings than another semantic-preserving embedding alternatives, such as Word2vec [38] (see Section 4.2.1 for more embedding alternatives).

These two vector encodings (node type vector and node code vector) are then concatenated to form the node representation vector. More concretely, considering the $i^{th}$ node of the $n^{th}$ graph, we denote its type as $t_{n,i} \in \{1,2,\ldots,69\}$, its code as $c_{n,i} \in \{1,2,\ldots,D\}^{l_{n,i}}$ where $D$ denotes the size of the dictionary and $l_{n,i}$ denotes the number of tokens of the node code $c_{n,i}$. The resulting node vector is denoted as $x_{n,i}$, that is obtained through the following process, where we use $[\cdot]$ to denote concatenation:

$$x_{n,i} = [e^{(t_{n,i})}, \text{Transformer}(c_{n,i})]$$

**Edge Vectors:** Joern's conversion of source code to graph yields 12 different types of edges, for example *USE, DEF, CONTROLS*, etc. To this set, we add the Next Consecutive Sequence (NCS) edges of the AST leaf nodes to maintain the sequential information in the graph [11]. Each edge is then encoded as a 13-dimensional one-hot vector. Formally, we denote the edge vector between the $i^{th}$ node and the $j^{th}$ node of the $n^{th}$ graph as $e^{n,i,j} \in \{0,1\}^{13}$.

### 3.3 Information Exchange

This step takes the information encoded in the nodes at the local statement-level from step 3.2, and performs a global information exchange throughout the source code graph. This is done for each graph in the pipeline ({AST, CFG, DFG}) separately. The idea is to gather information around each node's neighborhood, weighted according to the different edge types. After this information exchange, a graph-level representation is formed from the information accumulated at the relevant nodes. Finally, graph-level representations from the three pipelines are aggregated, and sent to the next (classification) layer.

---

[2] An encoding scheme to convert categorical features into a numeric array. Each category gets encoded in a num_categories dimension array, with a '1' only at the location of the corresponding category, and '0's elsewhere.

**Graph Convolution:** The core advantage of our approach comes from the disaggregated treatment of the source code graph. This, combined with a pooling module (described next) which drops irrelevant nodes, allows our model to focus on the nodes and edges better representative of vulnerabilities. As such, it is independent of the module responsible for message massing, with available alternatives including R-GCN [50], CGCN [60], GGNN [33], for example. We chose the crystal graphical convolutional neural network (CGCN) [60] for aggregating neighbouring nodes and edges' information during message passing. Our rationale behind this decision is two-fold: First convolutional graph neural networks (such as CGCN) are more scalable than recurrent graph neural networks (such as GGNN) [59], since the latter needs to compute and store several states for one node at the same time, while in CGCN each node vector is computed and updated immediately once per-layer. Second, CGCN is more efficient compared to other convolutional graph neural networks which also take in edge information, such as R-GCN. Unlike the latter, which computes update from the neighbouring nodes repetitively for each type of connections, CGCN takes only one aggregation step for all types of neighbouring edges and nodes.

The actual information exchange between the nodes of each code graph is governed by the following equation:

$$x_{n,i}^{'} = x_{n,i} + \sum_{j \in N(i)} \sigma([x_{n,i}, x_{n,j}, e_{n,i,j}]W_1 + b_1) \\ \odot g([x_{n,i}, x_{n,j}, e_{n,i,j}]W_2 + b_2)$$

where $x_{n,i}^{'}$ and $x_{n,i}$ denote updated and original $i^{th}$ node's vector of the $n^{th}$, $N_i$ denotes the neighbours of the $i^{th}$ node respectively, $\sigma$ denotes the Sigmoid activation [41] and $g$ denotes the Softplus activation [41]. $W_1, W_2, b_1, b_2$ are learnable weights.

**Hierarchical Modules:** Vulnerability analysis requires a model to produce graph-level predictions. However, typical vulnerabilities such as buffer overflow or pointer dereference usually would only take place in few statements, thus the relevant nodes are often the minority and may not be directly connected to each other. Instead of just one message passing layer, we use a hierarchical approach to filter out irrelevant nodes, to allow the our model to focus more on the set of nodes and edges better representative of vulnerabilities.

Specifically, at each layer, node vectors are updated by a CGCN module, and are then passed to a Self Attention Graph (SAG) Polling module [31] which will drop half of the nodes in the current graph. The rest of the nodes will be passed to the next layer. We read out the graphical representations at the end of each layer by a soft attention module, and sum all the layer-wise representations together to form the final graph representation (per pipeline).

The precise steps to narrow down on the relevant nodes and edges is as follows. The steps are the same for each pipleline.

1. At each layer (L), the node representations are updated according to the aforementioned CGCN module.

$$x_{n,i}^{(L+1)} = \tilde{x}_{n,i}^{(L)} + \sum_{j \in N(i)} \sigma([\tilde{x}_{n,i}^{(L)}, \tilde{x}_{n,j}^{(L)}, e_{n,i,j}]W_1^{(L)} + b_1^{(L)})$$
$$\odot g([\tilde{x}_{n,i}^{(L)}, \tilde{x}_{n,j}^{(L)}, e_{n,i,j}]W_2^{(L)} + b_2^{(L)})$$

2. Then we select a subset of the nodes to the next layer by SAG Pooling, in which the attention scores are calculated via a vanilla graph convolutional operator:

$$\alpha_{n,i}^{(L+1)} = \theta_1^{(L)} x_{n,i}^{(L)} + \sum_{j \in N(i)} \theta_2^{(L)} x_{n,j}^{(L)}$$

where $\theta_1^{(L)}, \theta_2^{(L)}$ are learnable weights.

3. We apply the attention score to the nodes in the top-k subset, while the other nodes and their edges are dropped from the current graph:

$$\tilde{x}_{n,i}^{(L+1)} = x_{n,i}^{(L+1)} \odot \tanh(\alpha_{n,i}^{(L+1)}) \quad \text{if } i \in \text{top}_k(\alpha_{n,i}^{(L+1)})$$

4. Graph-level soft-attention read-out is performed at the end of each layer, as follows:

$$O_n^{(L+1)} = \sum_{i \in V_n} \text{Softmax}[\text{MLP}_1(\tilde{x}_{n,i}^{(L+1)})] \odot \text{MLP}_2(\tilde{x}_{n,i}^{(L+1)})$$

where $\text{MLP}_1$ and $\text{MLP}_2$ denote two distinct multi-layer perceptrons [44] with learnable weights and standard ReLU activation [26].

5. Finally, the graph level read-outs are summed together to get the final graph representation:

$$g = \sum_L O_n^{(L)}$$

### 3.4 Classification Layer

In order to provide the class predictions, the final graph representations across the three pipelines are concatenated and then fed to a classification MLP, as follows:

$$\hat{y}_n = \sigma(\text{MLP}([g_{\text{AST}}, g_{\text{DFG}}, g_{\text{CFG}}]))$$

where $\sigma$ denotes the Sigmoid activation, and $g_{\text{AST}}, g_{\text{DFG}}, g_{\text{CFG}}$ represent the graph representations for AST, DFG and CFG respectively. Essentially the model assigns probabilities to the sample of belonging to different classes, based on the sample possessing the characteristics of a healthy vs. vulnerable code. These characteristics or signals is what the model learns during training as it gets exposed to multiple examples.

**Loss Function:** For vulnerability analysis datasets with binary labels, Cross Entropy Loss [24] with re-balancing class weights is often sufficient to provide good results. In case of multi-label datasets such as Draper [36], this is not sufficient, due to the highly unbalanced label distribution and the possibility that one function could fall into multiple vulnerability classes. We designed a novel loss function for such scenario that combines the Multi-class Binary Cross-Entropy (MBCE) loss and the Binary loss that only takes in the most confident prediction score. For a given ground truth $y \in \{0,1\}^M$ and prediction $\hat{y} \in [0,1]^M$ pair, the proposed loss function is computed as follows:

$$\text{MBCELoss}(y, \hat{y}) = -\frac{1}{M} \sum_{i=1}^{M} y_i \cdot \log \hat{y}_i + (1 - y_i) \cdot \log(1 - \hat{y}_i)$$

$$\text{BinaryLoss}(y, \hat{y}) = \begin{cases} -y_i \cdot \log \hat{y}_i - (1-y_i) \cdot \log(1-\hat{y}_i), \\ \quad i = \arg\max_i \hat{y}_i; \quad \text{if } \max(y) = 0 \\ -y_i \cdot \log \hat{y}_i - (1-y_i) \cdot \log(1-\hat{y}_i), \\ \quad i = \arg\max_i \hat{y}_i * y_i; \quad \text{if } \max(y) > 0 \end{cases}$$

$$\text{Loss}(y, \hat{y}) = w_1 \cdot \text{MBCELoss}(y, \hat{y}) + w_2 \cdot \text{BinaryLoss}(y, \hat{y})$$

where $w_1, w_2$ denote the balancing coefficients for the two components, and $M$ denotes the number of possible classes, in the case of Draper [36] $M = 5$. Note that just as normal cross entropy loss functions, class weights can be easily added to adjust for imbalanced classes.

The two-factor loss function has a number of appealing features for multi-class multi-label vulnerability detection. First, it addresses the binary classification and multi-class multi-label classification at the same time, thus spares the effort to train two classification modules respectively. Second, it offers the option to weigh more binary accuracy than getting all the class labels correct. This option is relevant to our application domain since vulnerability detection itself is a fairly difficult problem, and identifying the right class of vulnerabilities would be arguably more difficult. Being able to point out more vulnerabilities with some class missing labels is typically better than pointing out less vulnerabilities with their correct class labels. Third, from our empirical evaluations we found out that training with such fused targets gave superior performance than training with each target separately.

## 4 Evaluation

We first describe the datasets used for vulnerability detection. We then present a qualitative comparison of the different alternatives to vectorize source code. Next, we describe the baseline models we compare our 3GNN model against, and the model configurations. Finally, we present the results of the vulnerability prediction performance comparison for the the different models.

**Table 1:** Datasets overview.

|  | Draper | QEMU+FFmpeg |
|---|---|---|
| Total Samples | 1,192,509 | 21,020 |
| Vulnerabilities | 76,188 | 9,116 |
| Non-Vulnerabilities | 1,116,321 | 11,904 |
| Avg. No. of Nodes | 162 | 48 |
| Max. No. of Nodes | 500 | 355 |

## 4.1 Datasets

**Draper** [36] is a large and highly imbalanced dataset with over a million C/C++ functions collected from open source repositories on Github, and the Debian Linux distribution. Each function was checked by three static analyzers and labeled by security experts. Note that the quality of the labels strongly depends on the static analyzers—an imperfect but acceptable approximation.

**QEMU+FFmpeg** [62] is a much smaller but balanced dataset collected from Github repositories. The labelling is done based on commit messages and domain experts.

For both datasets, we removed the functions which Joern could not parse correctly, and those with more than 500 nodes due to memory limits. For Draper, we use the 80/10/10 split provided by [36], for QEMU+FFmpeg, we use 3-fold cross validation and a 75/25 split as suggested in [62]. The datasets are summarized in Table 1.

## 4.2 Baselines

Here we describe the multiple source code embedding schemes we experimented with, as well as various text, image and graph baseline models we compare 3GNN with.

### 4.2.1 Embedding Alternatives

We experimented with the following source code embedding schemes:

1. **Random embeddings** atop a dictionary of tokens extracted from source code across the dataset

2. **Normalizing** all identifiers to a generic `ID` token to avoid learning an incorrect association of identifiers (which change across samples) with vulnerabilities

3. **Symbolically normalizing** identifiers as per their appearance order in each sample, e.g. `VARIABLE_0`, `FUNCTION_1`, etc.

4. Vectorizing tokens using **Word2vec** [38] to preserve the semantic attributes of the code and language. For example, the vectors assigned to say `memcpy` and `memset` will be closer to each other, and distant to those for digits.

In our preliminary experiments we observed that ID normalization did not show any significant performance improvement over dictionary-based embeddings. In fact, it performed worse for non-graph-based models, because they did not have any auxiliary signals (like a graph's edge dependency) to compensate for the lost identifier context. Instead, such models performed better with the less-extreme symbolic normalization which was able to preserve enough signals, while reducing identifier-induced noise. Finally, the choice of whether or not to use Word2vec was highly dependent on the dataset and the model being trained. In light of these observations, we use the best performing encoding scheme for evaluating the baseline models described next.

### 4.2.2 Models

We use multiple state-of-the-art neural network baselines to compare 3GNN against. These operate upon different representations of source code: code-as-sequence, code-as-photo, and code-as-graph. These have already been shown to be superior to traditional approaches, as well as classical machine learning approaches in prior work.

**BiLSTM** is a kind of recurrent neural network (RNN). It processes source code as a sequence of tokens and tries to extract temporal signals. It has been shown to have good vulnerability detection performance [35]. BiLSTM performed best with Word2Vec embeddings for QEMU+FFmpeg dataset. No extra benefits were observed for Draper.

**CNN** treats source code as images, and then try to extract *pixel signals* from it using Convolutional Neural Network models from the image domain. This model, borrowed from [45], is a standard 1D-convolutional network and uses random token embeddings.

**GGNN**: Instead of borrowing techniques from image and time-series domain, this model operates at a more natural graph-level representation of source code. Amongst the existing graph neural network alternatives, we choose GGNN [33] as a comparison baseline since it has been shown to be quite effective for source code understanding tasks [11]. It uses message passing for transferring information from adjacent nodes over the connecting edges, and aggregating such information at each node via Gated Recurrent Units, a kind of RNN. For this model, the same node encoder as for our 3GNN model performed the best. There is one more recent work [62] which is a slight variant of the vanilla GGNN mentioned above, adding a convolution layer after the GGNN. Reproducibility and correspondence issues (discussed in Section 5) prevents its inclusion into the comparison baselines.

## 4.3 Model Configurations and Training

We use the hyper-parameters settings as suggested in [45, 62] for the baseline models. On top of the already extensively tuned hyper-parameters, we performed further tuning and confirmed the goodness of the configurations. For our model, we tuned hyper-parameters such as dropout, text embedding and hidden dimensions. We set the class weights according to the inverse class ratio for each dataset. For the transformer encoder, one layer of transformer is used with 2 attention heads and 64 hidden forward dimensions. Three attention read-out heads are used with Softmax activation at the end of the node encoding module for Draper while one attention head is used for QEMU+FFmpeg due to its smaller data size. As for the CGCN convolutional module, input and output dimensions are set to the number of node types plus the dimension of transformer encoder output. All of the intermediate SAG Pooling modules are configured to drop 50% of the nodes upon taking inputs; they keep the output dimensions of the same size as the input. In the end, the three 128-dimensional vectors representing each sub-graph are concatenated and fed into the MLP to generate the final predictions.

Table 2: Comparison of our model 3GNN with baselines for Draper and QEMU+FFmpeg datasets.

|  | Draper | | | QEMU+FFmpeg | | |
|---|---|---|---|---|---|---|
| Model | Acc | F1 | MCC | Acc | F1 | MCC |
| BiLSTM | 92.04 | 49.35 | 0.461 | 60.32 | 61.71 | 0.243 |
| CNN | 92.26 | 49.40 | 0.460 | 59.01 | 61.36 | 0.227 |
| GGNN | **93.49** | 50.80 | 0.474 | **61.86** | 59.0 | 0.239 |
| 3GNN | 93.26 | **54.32** | **0.512** | 61.03 | **61.83** | **0.252** |

Table 3: Extra performance squeezed by adding a Random Forest classifier on the final feature representation generated by the models just prior to the Classification layer. Results for the Draper dataset. Improvement observed irrespective of the models' treatment of code as linear sequence, or code as photo, or code as graph.

| Model | Acc | F1 | MCC |
|---|---|---|---|
| BiLSTM+RF | 92.79 | 51.13 | 0.477 |
| CNN+RF | 92.70 | 52.72 | 0.497 |
| 3GNN+RF | **93.22** | **54.94** | **0.520** |

In the Draper experiments, we trained for 100 epochs and evaluated performance on the test set according to the best validation F1 score. For QEMU+FFmpeg we used an early-stop mechanism with patience of 100 epochs (as [62]), and we report the averaged cross-validation performances.

## 4.4 Results and Discussion

Table 2 summarizes the results on both datasets. Shown are the accuracy, the F1 score, and the Matthews correlation coefficient (MCC). The classes in Draper are highly imbalanced and thus the last two metrics more reliably measure the performance [13, 16].

As can be seen, 3GNN outperforms all baselines. Being able to extract 6.9% more F1 performance from even a highly imbalanced dataset as Draper, shows the ability of our model to enhance a generic graph neural network so as to better capture vulnerability signals. Considering that the node encoder is the same for both GGNN and 3GNN, this performance improvement is attributable to the independent use of the sub-graphs combined with the hierarchical modules, allowing our model to better localize attention towards potential vulnerability templates.

In case of QEMU+FFmpeg, the margin is not that significant. A possible reason is the limited size of this dataset, which makes it hard to differentiate between the potential of each model. This is corroborated by the significant over-fitting we saw for all models on this dataset, even with heavy regularization. Still, the performance relative to the Draper dataset is better, indicating potential for practical use.

To measure the contribution of each individual graph component, we evaluate the prediction performance achieved by our model using each graph in isolation- AST vs. CFG vs. DFG. Learning based only on the CFG representation of source code leads to a 37% drop in F1, compared to incorporating information from all three pipelines in our 3GNN model. DFG alone fairs better with a 14% decline in F1. AST provides the most relevant information coming close to within 3.5% of 3GNN's F1.

The large size of the Draper dataset enables stacking more layers to squeeze some more performance from the models [45]. For example, as shown in Table 3, adding a Random Forest layer [14] on top of the models improves model performance across the board, by up to 6.7% as observed for the CNN model. The absolute numbers are still low, as opposed to the model performance (> 0.8 F1) seen on synthetic datasets in previous work [45, 52], highlighting the complexity of a noisy real-world dataset. Another reason for the lower performance on Draper could be the quality of labels, having been derived from static analyzers. A potentially erroneous ground truth can greatly affect the ability of the models to learn the differentiating properties of vulnerable vs. healthy code.

## 5 Related Work

Recently, several ML models have been proposed to help alleviate some of the issues of traditional techniques, as discussed in Section 2. These methods tend to get signals from source code features such as number of lines or conditional statements, use of sensitive library functions or system calls, call-stack depth, complexity measures like cyclomatic or Halstead complexity, etc. [9, 23, 25, 37, 47, 53]. This can be augmented with meta features such as commit messages and bug reports [46, 63].

Another approach to automated source code understanding is via statistical language models. These methods capture regularities in source code [28] at the token level. N-gram approaches and Hidden Markov Models have been used for tasks such as code completion, fault localization, and code search [30, 39, 43]. Specific to the bug detection task, [57] leverages n-gram language models to calculate the probability distribution of program tokens in a project, and flag low probability token sequences as potential bugs. More recently, deep learning models have emerged as same or better performers than n-gram models [58].

Instead of feature engineering as in classical ML, deep learning approaches automatically extract signals from code by treating it as an image or a linear sequence [17, 27, 34, 35, 42, 45]. We believe more meaningful signals are preserved in a graph-level encoding of source code and thus experiment with a code-as-a-graph representation. We are motivated by the template matching work (non-learning-based) introduced by Yamaguchi et al. [61]. Unlike their work which uses a multi-graph, we process the individual graph components in separate pipelines for more fine-grained signal extraction.

From an encoding perspective, recent works have also looked into deep learning over code-as-graph as in [11, 12, 54, 56]. [56] uses graph neural network based auto-encoders for unsupervised learning over source code ASTs to automatically cluster Java classes into different categories (business logic, interface, and utility classes). [11] uses Gated Graph Neural Networks to learn fundamental program structures, such as predicting variable names given its usage, or selecting the correct variable given a program location. Similarly, [54] uses GNN to learn loop invariants for program verification. Code2vec [12] uses a collection of path-contexts (AST paths + leaf-node values) to encode source code, and an attention-based neural network to learn appropriate names for a function from its code. In our case, we operate at a macroscopic level of automatically learning vulnerability templates from source code graphs.

Most relevant work to ours which also uses deep learning over code-as-graph in the software vulnerability domain includes [19, 32, 62]. [32] uses a combination of AST, PDG and DFG graphs to encode Java methods. Attention GRU as well as attention convolutional layer is used to focus on buggy paths in the code. A direct comparison is not straightforward due to a difference in the target language (the Joern graph extractor we use is limited to C/C++) and the type of software bugs in focus (we don't target named-based bugs). The contrast is similar with Hoppity [19], which adds a pointer mechanism to a GNN [49] for bug localization. However, unlike 3GNN's focus and disaggregated graph methodology, it targets general programming bugs specific to Javascript code, and operates primarily on the AST. Devign [62] comes closest to our approach, using GNNs and operating upon Joern graphs on C/C++ code. It uses a multi-graph representation tying different source code graphs together, as opposed to our philosophy of assisting the model in concentrating on specific program constructs independently. While we train on all Joern edges, Devign uses a subset of it's edges, while adding a few other edge types inspired by [11]. The Devign paper does not present results on already existing datasets but introduces it's own dataset, half of which has been publicly released. We do not use this method as a baseline for comparison because we are not convinced about our model reproduction based on the paper details. We were unable to achieve the results presented in the paper despite our best efforts. Several attempts of correspondence with the authors for clarifications also failed.

Finally, there is yet another alternative of treating code-as-natural-language, where approaches such as BERT [18], which have shown great promise in the Natural Language Processing (NLP) domain, can be put to use. Recently, it has been adopted for source code understanding to perform tasks such as variable misuse prediction, incorrect operator/operand detection, code search, documentation generation, and function-docstring mismatch [22, 29]. We are still exploring this dimension of code representation for vulnerability detection, and do not compare with it in this work.

## 6 Acknowledgement

The authors would like to thank Dr. Jie Chen (MIT-IBM Watson AI Lab, IBM Research) for his guidance in this work.

## 7 Conclusion

We presented a novel graph neural network architecture for automatically detecting vulnerabilities in source code. We enhance a generic GNN with the ability to better localize attention towards potential vulnerability templates by (i) using multiple graph representations of a program separately, (ii) employing hierarchical layers of convolutional and pooling modules, enabling the network to drop irrelevant nodes, and (iii) developing a new training loss metric tailored for vulnerability analysis type tasks with multi-class / multi-labels. Our model outperforms several text (BiLSTM), image (CNN) and graph (GGNN) baseline models, recording a 6.9% better F1 than the next-best model (GGNN) for the real-world Draper dataset. Future work includes the investigation of the explainability for this model, and its application to new vulnerability detection datasets.